\documentclass[letterpaper,twocolumn,10pt]{article}
\usepackage{zhanggroup}

\usepackage{xspace}
\usepackage{amsmath,amssymb,amsfonts}
\usepackage{algorithmic}
\usepackage{graphicx}
\usepackage{textcomp}
\usepackage{xcolor}
\usepackage{tikz}
\usepackage{array}
\usepackage{pifont}
\usepackage[normalem]{ulem}
\usepackage{multirow}
\usepackage{subcaption}
\usepackage{booktabs}
\usepackage{makecell}
\usepackage{url}
\usepackage{tcolorbox}
\usepackage{hyperref}
\hypersetup{
  colorlinks,
  linkcolor={blue!70!green},
  citecolor={green!70!blue},
  urlcolor={orange!70!red}
}
\graphicspath{ {images/} }

\newcommand{\mypara}[1]{\smallskip\noindent{\bf {#1}.}\xspace}

\begin{document}

\title{Assessing Deanonymization Risks with Stylometry-Assisted LLM Agent}

\date{}

\author{
 \textbf{Boyang Zhang},
 \textbf{Yang Zhang}
\\
\\
 CISPA Helmholtz Center for Information Security
\\
}

\maketitle

\begin{abstract}

The rapid advancement of large language models (LLMs) has enabled powerful authorship inference capabilities, raising growing concerns about unintended deanonymization risks in textual data such as news articles.
In this work, we introduce an LLM agent designed to evaluate and mitigate such risks through a structured, interpretable pipeline.
Central to our framework is the proposed $\textit{SALA}$ (Stylometry-Assisted LLM Analysis) method, which integrates quantitative stylometric features with LLM reasoning for robust and transparent authorship attribution.
Experiments on large-scale news datasets demonstrate that $\textit{SALA}$, particularly when augmented with a database module, achieves high inference accuracy in various scenarios.
Finally, we propose a guided recomposition strategy that leverages the agent's reasoning trace to generate rewriting prompts, effectively reducing authorship identifiability while preserving textual meaning.
Our findings highlight both the deanonymization potential of LLM agents and the importance of interpretable, proactive defenses for safeguarding author privacy.

\end{abstract}

\section{Introduction}
\label{section:introduction}

Recent advancements in large language models (LLMs) have enabled the development of increasingly capable AI agents that integrate LLM reasoning with tool use, autonomous planning, and multi-step execution. 
Modern LLM-based agents can not only generate fluent and contextually relevant text but also interact with external environments.
Such capabilities have transformed AI agents from passive text generators into active problem solvers, deployed across domains including research assistance, software engineering, and content moderation~\cite{XCGHDHZWJZZFWXZWJZLYDWCZQZQHG23,JWMLW24}.

However, as LLM agents grow more powerful, their autonomy and reach raise serious safety and ethical concerns.
Unlike static models, agents with access to external tools and dynamic reasoning capabilities can produce outputs with real-world consequences. 
These capabilities expose a new vector of privacy risk, particularly for individuals who rely on anonymity or pseudonymity in digital communication.
For instance, whistleblowers, journalists, and researchers in sensitive areas often depend on concealing authorship to ensure personal safety or maintain impartiality.
Authorship attribution has been studied extensively in computational linguistics.
Traditional approaches rely on supervised models trained on handcrafted features (e.g., word frequency, function word distribution, syntactic structure)~\cite{NPGBSSS12,BRMS23}. Recent advances have shown that LLMs can implicitly encode authorial style, enabling zero-shot or few-shot authorship identification without explicit training~\cite{HCS24}.
Now with tool-augmented LLM agents, these capabilities can even be further amplified.

Motivated by this growing threat to textual anonymity, in this work, we introduce an LLM agent framework designed to evaluate and mitigate deanonymization risks in news articles, a particularly fitting target for the subject.
Our agent operates through a four-stage pipeline including information extraction, candidate search, candidate matching, and result analysis. We propose a novel matching strategy, \textsc{SALA} (Stylometry-Assisted LLM Analysis), which integrates stylometric feature evaluation with LLM reasoning to achieve precise and explainable authorship attribution. To further enhance scalability, we implement a database module that aggregates author profiles and sample writings, significantly improving both search efficiency and matching accuracy.
Finally, we introduce a guided recomposition strategy that leverages the agent's reasoning trace to provide targeted rewriting prompts, effectively reducing authorship identifiability while preserving content utility. 
For example, if the agent determines that an author is identifiable due to consistent use of domain-specific terminology or idiosyncratic sentence structure, it can recommend lexical diversification, style normalization, or semantic rephrasing strategies.
In doing so, the agent serves as both an adversary and advisor, quantifying authorship risk and guiding authors toward safer textual representations.
Together, these components establish a unified framework for understanding and mitigating LLM-agent-driven deanonymization risks.

In summary, this work contributes:
\begin{enumerate}
    \item An LLM-based multi-tool authorship inference agent, capable of identifying potential authors through stylistic and semantic analysis augmented with external retrieval.
    \item A privacy evaluation framework that quantifies the strength of authorship signals and explains how they were derived.
    \item An anonymity enhancement module that provides evidence-based recommendations to reduce identifiability.
\end{enumerate}

By simulating a powerful adversarial agent and offering defensive interventions, our approach bridges authorship attribution research with privacy-aware text generation, addressing an increasingly urgent challenge in the era of autonomous AI systems.

\section{Background}
\label{section:background}

\subsection{LLM Agent}
\label{subsection:background_agent}

LLM agents are typically composed of four key components: \textit{core}, \textit{planning}, \textit{tools}, and \textit{memory}~\cite{LYZXLLGDMYZDZDZSZSSHDT23,RDWPZBDMH24}.
These components extend LLMs' capabilities beyond simple text generation to executable actions in both the digital and physical worlds.
While these extended capabilities have enabled promising applications across diverse domains, they have also raised growing concerns regarding potential risks and unintended misuse~\cite{GZPDLWJL24,RDWPZBDMH24,ZLYK24,YBLCZS24,MLZSXS24,ZTSSBZZ25,DZBBFT24}.

\subsection{Deanonymization}
\label{subsection:background_anonym}

Authorship attribution (or author identification) is the task of determining the most likely author of a text from a candidate set.
More broadly, authorship analysis also includes author profiling (inferring demographic or stylistic attributes) and authorship verification (deciding whether a text was written by a given author) \cite{HLVS24}.
Modern approaches leveraging machine-learning-based methods have also achieved success~\cite{NPGBSSS12,BRMS23}. Recently, the introduction of powerful LLMs presents new vectors where the risk of unintentional deanonymization can arise~\cite{HCS24}.
Similar analysis has also been adapted in distinguishing LLM versus human-generated texts~\cite{MSS23,O24,KGBTRL23}.

\section{Authorship Agent Design}
\label{section:agent_design}

Our proposed deanonymization risk assessment agent is designed as a modular system that integrates reasoning capabilities of LLMs with external tools for information retrieval, stylometric analysis, and interpretability.
The agent operates through four sequential stages: \textit{Information Extraction}, \textit{Candidate Search}, \textit{Candidate Matching}, and \textit{Result Reflection}.
Additionally, the agent also integrates a \textit{Database Module} for improving inference performance and a \textit{Anonymization Enhancement Module} for mitigating the discovered risk.

\subsection{Stage: Information Extraction}
\label{subsection:information_extraction}

In the first stage, the agent analyzes the given article to extract relevant metadata.
Metadata includes information such as publication date, topic category, and geo-location.
This step is to obtain as much information as possible to find the subsequent direction of finding potential author candidates.
This information can be extracted easily utilizing LLMs, using prompts such as ``Extract the main topic category, publication date, and potential publisher origins.''

\subsection{Stage: Candidate Search}
\label{subsection:candidate_search}

Next, the agent performs a candidate search to retrieve a set of potential authors.
By default, this stage is powered by a web search function following ChatGPT's API specification.
The function queries publicly available sources for author-article associations. 
The candidate search is semi-guided with instructions that draw the LLM's ``attention'' towards the key information extracted from the previous stage.
Users will also specify the number of potential candidates to collect.

After the list of potential candidates is determined, the agent will continue to search for and collect sample articles authored by the candidates.
Users will also specify the number of samples per candidate to collect.

\subsection{Stage: Candidate Matching}
\label{subsection:candidate_matching}

\begin{table*}[!t]
\centering
\caption{Stylometric features used in constructing our Stylometry-Assisted LLM Analysis (SALA).}
\label{table:stylo}
\scalebox{1.0}{
\begin{tabular}{@{}l|l|l@{}}
\toprule
\textbf{Category}          & \textbf{Feature}          & \textbf{Description}                                                      \\ \midrule
\multirow{4}{*}{Lexical}   & Unique word count         & Total number of distinct words used in the text.                          \\
                           & Average word length       & Mean number of characters per word.                                       \\
                           & Type-token ratio          & TTR: ratio of unique words to total words.                                \\
                           & Hapax Legomenon ratio     & Proportion of words that appear only once in the text.                    \\ \midrule
\multirow{4}{*}{Syntactic} & Average sentence length   & Mean number of words per sentence.                                        \\
                           & Stopword count            & Number of common functional words (e.g., "the," "is") used.               \\
                           & Punctuation count         & Total occurrences of punctuation marks.                                   \\
                           & POS variation count       & Diversity of parts of speech (e.g., nouns, verbs, adjectives) used.       \\ \midrule
Readability                & Flesch-reading-ease score & A readability score based on sentence length and syllable count.          \\ \midrule
\multirow{2}{*}{Semantic}  & Polarity                  & Measuring whether overall tone is positive or negative.                   \\
                           & Subjectivity              & Degree to which the text is subjective opinion or objective facts.        \\ \midrule
Style                      & Writing style summary     & Qualitative style analysis, e.g., formal, casual, with humor/sarcasm, etc. \\ \bottomrule
\end{tabular}
}
\end{table*}

Once the candidate set is established, the agent conducts detailed pairwise matching between the target article and each candidate's author profile.
By default, the input article and every sample collected from the previous stage are compared using a specified matching strategy.

Here, we propose a new authorship matching method, \textbf{Stylometry-Assisted LLM Analysis (SALA)}, which integrates quantitative stylometric features with LLM-based analysis to determine authorship attribution.
Specifically, we extract a suite of stylometric features from both the target and candidate articles, as summarized in \autoref{table:stylo}.
The selected features span five categories: \textit{lexical}, \textit{syntactic}, \textit{readability}, \textit{semantic}, and overall \textit{style}.
Features in the first three categories can be computed directly from the text using Python-based NLP tools (e.g., \textit{NLTK}), while the \textit{semantic} and \textit{style} features are obtained by querying the LLM itself.
This approach avoids the need to train additional models and enables dynamic extraction of higher-level stylistic cues.

The extracted features are combined into a structured description, which is then used to guide the LLM's pairwise comparison.
The agent is prompted to assess authorship similarity given these feature values. For instance, a typical prompt might be:  
\textit{``Given Article A has a unique word count of $n_1$, average word length of $m_1$, etc., and Article B has a unique word count of $n_2$, average word length of $m_2$, etc., what is the likelihood that these two articles were written by the same author?''}  
This structured prompting strategy grounds the model's reasoning in explicit, interpretable metrics while retaining its ability to synthesize qualitative insights.

Both stylometric features and LLM-based analyses have been explored in prior authorship attribution research~\cite{NPGBSSS12,BRMS23}.
However, conventional stylometric approaches are predominantly training-based, using extracted features as input to supervised models, which often limits adaptability across domains.
Meanwhile, prior LLM-based approaches~\cite{HCS24} primarily rely on qualitative style descriptions (e.g., analyzing phrasal verbs, punctuation, or humor) without quantitative grounding, leading to less precise results, especially given that LLMs still suffer from issues such as hallucinations~\cite{LCZNW23,JLFYSXIBMF23,BCLDSWLJYCDXF23,RSD23}.

To contrast with our approach, we define this qualitative prompting baseline as \textbf{LLM Direct Analysis (LDA)}, which we include for comparison.
We also evaluate a third baseline based on cosine similarity between text embeddings (\textbf{ES}).
In summary, by leveraging the LLM's tool execution capabilities, \textsc{SALA} combines deterministic feature evaluation with reasoning-driven comparison, enhancing both the precision and interpretability of authorship inference without requiring additional model training.

\subsection{Stage: Result Reflection}
\label{subsection:result_reflection}

In the final stage, the agent interprets the matching results to assess deanonymization risk and identify key reasoning steps that lead to the conclusion.
By analyzing the most influential features contributing to authorship inference, the system identifies stylistic elements that may reveal author identity.
This reflective component transforms the agent from a purely diagnostic tool into a proactive anonymization assistant when incorporated with appropriate modules.

\subsection{Module: Database}
\label{subsection:database}

In both the candidate search and matching stages, the web search function is particularly resource-intensive and time-consuming.
For each candidate author, their associated writing samples were retrieved, and it is therefore logical to establish a database module that allows the framework to reuse previously gathered information.

With the database module implemented, we introduce an initial ``warm-up'' phase to preload the system with candidate authors and their representative writing samples. Once the database accumulates a sufficient repository, the search stage can operate more efficiently by retrieving candidates through embedding similarity and keyword matching, resorting to external web search only when suitable matches are unavailable.

The matching stage also benefits from the database integration. Since the database stores both authors and their sample writings, comprehensive author profiles can be constructed using aggregated stylometric features. By summarizing metrics across multiple documents, the system produces a more representative and stable feature profile for each author. During matching, the input article is then compared against these aggregated author profiles rather than individual samples, enhancing both robustness and efficiency.

\subsection{Module: Anonymization Enhancement}
\label{subsection:defense_module}

Building upon the result reflection stage, the \emph{Anonymization Enhancer Module} translates attribution insights into actionable privacy guidance.
We propose a \textbf{guided rewrite} defense mechanism to mitigate potential deanonymization risks.
By examining the stylistic and contextual features that most contributed to author identification, the module recommends targeted modifications designed to reduce re-identifiability.
Suggested strategies may include:
Adjusting or diversifying stylistic features (e.g., sentence length, syntactic variation).
Substituting signature lexical patterns or domain-specific/high-complexity phrases.
Modifying discourse structures or thematic emphases associated with the inferred author.
The module can directly provide a \emph{guided recompose} based on the suggestion, by modifying the suggestions into instructions in the rewriting prompt:``Rewrite this article, follow these suggestions closely.''
If using LLM-generated text is not applicable, these suggestions still provide guidance for manual editing to mitigate the potential risk of deanonymization.

\subsection{Implementation}
\label{subsection:implementation}

We implement the agent using the \textit{gpt-4.1-2025-04-14} model~\cite{GPT4.1} via the OpenAI Chat Completions API, integrated within the \textit{LangChain} agent framework~\cite{LangChain}. The stylometric feature extraction tools are developed in Python, utilizing open-source NLP libraries such as \textit{NLTK}~\footnote{\url{https://www.nltk.org/api/nltk.html}} for computing lexical, syntactic, and POS-based features. 
A \textit{web search} function is also implemented following ChatGPT's API specification, enabling real-time retrieval of external information when required.
For the following evaluations, we experiment and report results after five repetitions, with the average values reported.

\section{Deanonymization Risk Evaluation}
\label{section:result}

\subsection{Attack Scenarios}
\label{subsection:scenarios}

To evaluate the agent's capability to compromise author anonymity for a given article, we consider two attack scenarios that simulate real-world situations in which potential adversaries aim to uncover the identity of authors of anonymized articles.

\mypara{Targeted Attack}
In the \emph{targeted attack}, the adversary examines whether a given text sample is written by one or several specific authors whose identities are known to the adversary and targeted for the attack.
In this scenario, the agent can bypass the candidate selection stage, since the adversary already provides the targets.
To evaluate attack performance in this scenario, we construct test sets containing articles from both target authors and non-target authors in equal proportions.
We then evaluate whether the agent can identify which articles from the test set are written by the target authors.
For multiple targets, the test set still contains 50\% articles from unrelated authors, and attack performance is evaluated based on whether the agent can identify the \emph{specific} author of a given article (not whether it was written by one of the targeted authors).
Therefore, in this setting, as the number of targets increases, the attack difficulty increases substantially.
We use the F1-score as the measuring metric to take both precision and recall into consideration.

\mypara{Open-World Attack}
In the \emph{open-world attack}, we examine the agent's deanonymization capability in a more challenging environment.
The adversary aims to identify the author's identity without a specific target or prior knowledge, using only the given news article.
The candidate search stage is required in this setting to generate a list of potential candidates before analyzing whether there is a positive match.
The agent outputs a ranked short list of potential authors ordered by likelihood.
Using a test set consisting of articles from various authors, we evaluate performance based on whether the true author appears in the agent's top-1 or top-3 predictions.

\subsection{Evaluation Dataset}
\label{subsection:dataset}

To evaluate our proposed framework, we conduct experiments on two large-scale open-source datasets: \emph{All the News 2.0}\footnote{\url{https://huggingface.co/datasets/rjac/all-the-news-2-1-Component-one}.} and \emph{CrossNews}~\cite{MLKDCFRX25}.
\emph{All the News 2.0} is a comprehensive collection of news articles spanning multiple U.S. media outlets, topics, and time periods.
It contains over 2.5 million articles from hundreds of publications, encompassing a wide spectrum of journalistic styles and editorial practices.
Importantly, the dataset includes byline information for a large number of authors (more than thirteen thousand), allowing for fine-grained analysis of author-level stylistic patterns and cross-source variability.
\emph{CrossNews}, while sharing overlaps, offers more in-depth categorization across different news domains and writing contexts.
In addition to standard news reporting, it includes multiple modalities of publication from the same author, such as blog posts, tweets of journalists, along with their standard professional writings.

\subsection{Candidate Search Performance}
\label{subsection:search_result}

\begin{table}[!t]
\centering
\caption{Percentages of test cases with true authors in the selected potential candidate lists in zero-shot and database augmented (DB-AUG) setting.}
\label{table:search_result}
\scalebox{1.0}{
\begin{tabular}{@{}ccc@{}}
\toprule
\multicolumn{1}{l}{\textbf{Potential Candidates}} & \textbf{Zero-Shot} & \textbf{DB-AUG} \\ \midrule
20                                       & 1.2\%              & 78.1\%          \\
50                                       & 9.8\%              & 89.5\%          \\
100                                      & 21.1\%             & 93.1\%          \\ \bottomrule
\end{tabular}
}
\end{table}

\mypara{Zero-Shot Search}
Given the challenging scenario of open-world authorship attribution, the agent will first have to narrow down potential candidates.
Following the pipeline introduced in \autoref{section:agent_design}, after obtaining the article, the agent utilizes both the web search function and the LLM's internal knowledge to generate a list of potential candidates.
To ensure downstream analysis efficiency, we set the number of candidates to 20 by default.
We select a random subset of 1,000 articles from the dataset and examine the agent's performance on correctly including the true author in the candidate pool.

\autoref{table:search_result} shows the percentages of test cases in which the correct author is included within the preliminary list provided by the agent when it is instructed to provide 20, 50, and 100 potential candidates.
Overall, this direct approach to narrowing down candidates is not very effective.
In the default setting of 20 potential candidates, the agent succeeds on only 1.2\% of test cases.
When we increase the number of potential candidates to 50 or 100, prediction performance improves as expected.
However, with 50 or 100 potential candidates, the downstream analysis's computational cost becomes exponentially higher and subsequently impractical.

\mypara{Database-Augmented Search}
When we incorporate the database component of the agent and allow it to warm up with an initial collection of a large pool of potential authors, performance improves substantially.
For efficiency, we manually create the database with a subset of the dataset that is disjoint from all data used in the evaluation for this paper.
We also intentionally omit some authors to simulate real-world scenarios where the initial database does not include all possible authors.
Given the seamless integration of the database in the agent, the searching stage is much more efficient than starting from scratch in the zero-shot scenario.
Finding related authors is also much more accurate with keyword and embedding similarity search within the database, compared to utilizing the costly web search function.
This is directly reflected in the searching performance, where with 20 potential candidates, the candidate coverage rate increases to 78.1\% and reaches as high as 93.1\% when increasing the potential candidates to 100.

Since the database already includes previous articles, and analysis features have been generated accordingly, computational complexity increases only linearly with the number of candidates.
In the zero-shot setting, however, the computational requirement increases exponentially with the number of candidates since relevant samples must be searched, collected, and analyzed for each candidate.
As a result, the matching process is also more efficient and can thus accommodate an increased number of potential candidates.

\subsection{Zero-Shot Match Results}
\label{subsection:zeroshot_result}

\begin{table}[!t]
\centering
\caption{Zero-shot attack performance with different matching strategies in the targeted attack scenario with different numbers of samples per
candidate and the number of targets. ES = Embeddings Similarity, LDA = LLM Direct Analysis, SALA = Stylometry-Assisted LLM Analysis, No.= Number of.}
\label{table:zero_target}
\scalebox{1.0}{
\begin{tabular}{@{}c|c|ccc@{}}
\toprule
\textbf{No. Target} & \textbf{No. Sample} & \textbf{ES} & \textbf{LDA} & \textbf{SALA} \\ \midrule
\multirow{2}{*}{1}   & 5                   & 0.500       & 0.500        & 0.619         \\
                     & 10                  & 0.513       & 0.500        & 0.645         \\ \midrule
\multirow{2}{*}{3}   & 5                   & 0.167       & 0.168        & 0.244         \\
                     & 10                  & 0.172       & 0.166        & 0.291         \\ \bottomrule
\end{tabular}
}
\end{table}

We first analyze the agent's zero-shot candidate matching performance without utilizing the database module in the two attack scenarios.

\mypara{Targeted Attack}
\autoref{table:zero_target} shows the F1 score of the agent's prediction on whether the given article is written by the targeted author.
We examine the three different methods' performance with varying numbers of targeted authors and writing samples provided for each target.
We find that directly comparing embedding similarity is not very effective.
This is understandable considering that nuanced differences in writing style are highly unlikely to be reflected in the embeddings, which are more suitable for general topic or thematic comparison.
The LDA approach is also not effective.
Through analyzing the responses, we find that when conducting pairwise comparison, the LLM suffers from hallucination and consistently determines that the two articles being compared differ significantly in style, even when they are both news articles written in a very formal style.
Solely relying on the LLM to directly analyze these texts and generate descriptions can be unreliable.

With SALA, we achieve the best performance in this setting.
The quantitatively calculated stylometric statistics in our SALA matching strategy assist the LLM to generate more grounded analysis.
However, the overall performance is still not adequate, reaching only 0.645 in the best setting.
This is likely due to the limited samples used for matching, constrained by the high computation cost of the zero-shot setting.

\begin{table}[!t]
\centering
\caption{Zero-shot attack performance in an open-world setting.}
\label{table:zero_open}
\scalebox{1.0}{
\begin{tabular}{@{}llll@{}}
\toprule
              & \textbf{ES} & \textbf{LDA} & \textbf{SALA} \\ \midrule
\textbf{Full}      & 0.000 & 0.000 & 0.000 \\
\textbf{Filtered}  & 0.053 & 0.058 & 0.068 \\ \bottomrule
\end{tabular}
}
\end{table}

\mypara{Open-World Attack}
As seen from \autoref{table:search_result}, without the memory module, the potential candidate selection already has very limited success.
Making further analysis based on this limited success in candidate selection does not yield any successful predictions, regardless of the matching strategy.
\autoref{table:zero_open} shows that using the full pipeline (matching with the searched results), the agent cannot make any correct predictions.
To minimize the influence of the candidate selection stage and focus on cases where the true author is among the selected candidates, we filter out instances with no correct candidate and evaluate the matching performance in the zero-shot setting in isolation.
Given the correct potential candidates, we find that the agent is able to make correct predictions following the pipeline, but still with very limited success across all matching strategies.
For more in-depth analysis and comparison, we require the database module to improve prediction performance first.

\subsection{Database-Augmented Match Results}
\label{subsection:db_result}

The incorporation of a database into the agent not only improves candidate search efficacy and efficiency, as shown in \autoref{subsection:search_result}, but also improves downstream matching performance by allowing a significantly larger number of samples to be used for analysis and matching.
We now present the agent's authorship inference performance in the two scenarios with the database module implemented.

\begin{table}[!t]
\centering
\caption{Authorship inference performance with different matching strategies in targeted
attack scenario with different numbers of samples per
candidate and the number of targets.}
\label{table:db_target}
\scalebox{1.0}{
\begin{tabular}{@{}c|c|ccc@{}}
\toprule
\textbf{No. Target} & \textbf{No. Sample} & \textbf{ES} & \textbf{LDA} & \textbf{SALA} \\ \midrule
\multirow{3}{*}{1} & 10          & 0.500 & 0.654 & 0.674 \\
                   & 200         & 0.500 & 0.699 & 0.801 \\
                   & 500         & 0.500 & 0.681 & 0.827 \\ \midrule
\multirow{3}{*}{3} & 10          & 0.167 & 0.243 & 0.312 \\
                   & 200         & 0.167 & 0.223 & 0.551 \\
                   & 500         & 0.167 & 0.198 & 0.595 \\ \bottomrule
\end{tabular}
}
\end{table}

\begin{figure}[!t]
\centering
\includegraphics[width=\columnwidth]{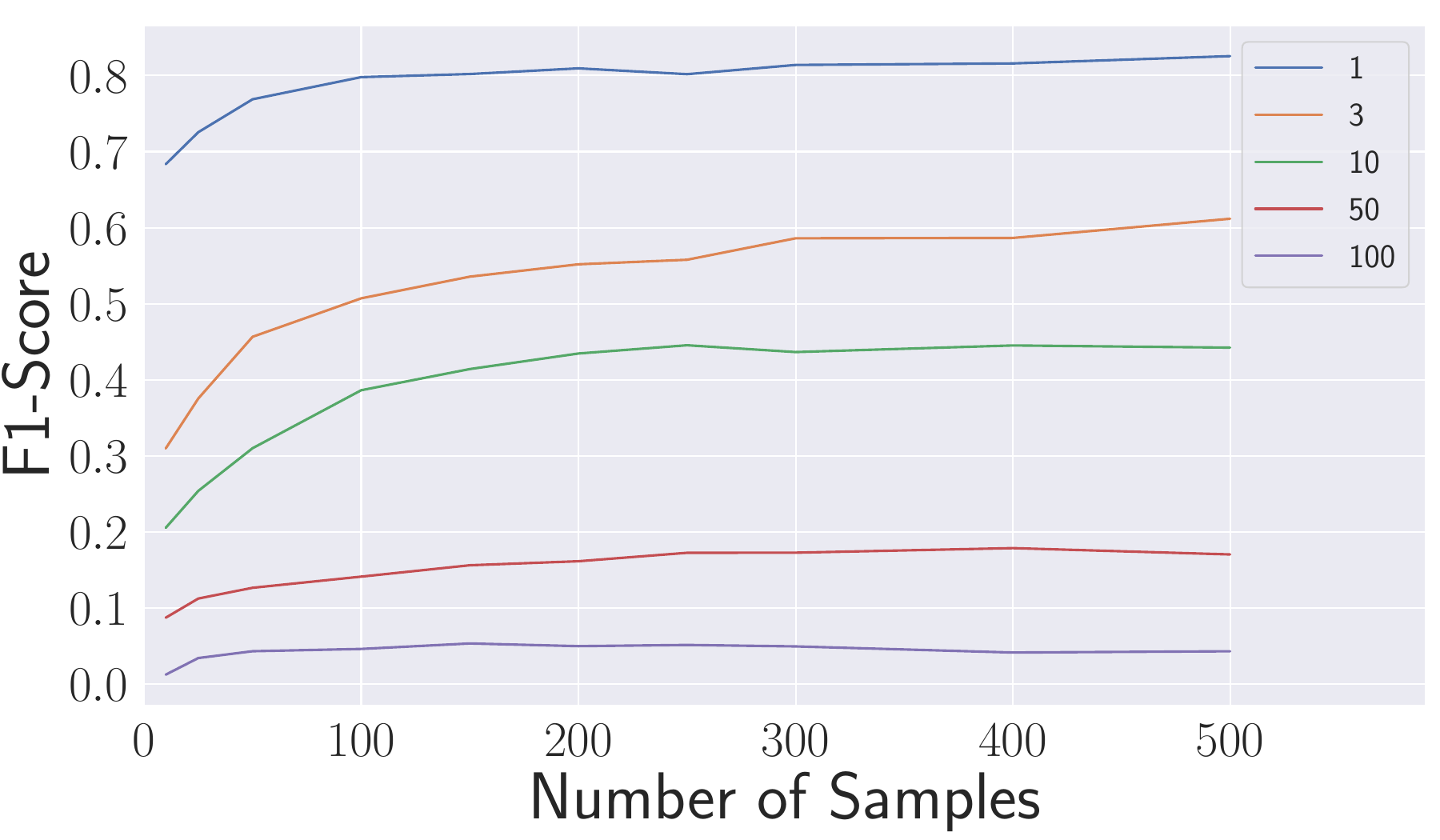}
\caption{Authorship inference performance in targeted attack scenario with different numbers of samples per candidate and number of targets.}
\label{figure:targeted}
\end{figure}

\begin{table}[!t]
\centering
\caption{Authorship inference performance when using only one type of stylometric feature in SALA.}
\label{table:db_sala}
\scalebox{1.0}{
\begin{tabular}{@{}l|ccc@{}}
\toprule
\textbf{Number of Targets}& \textbf{1}    & \textbf{3}    & \textbf{10}    \\ \midrule
\textbf{Lexical}     & 0.728         & 0.524         & 0.298          \\
\textbf{Syntactic}   & 0.62          & 0.446         & 0.308          \\
\textbf{Readability} & 0.545         & 0.199         & 0.120          \\
\textbf{Semantic}    & 0.579         & 0.245         & 0.089          \\
\textbf{Style}       & 0.654         & 0.188         & 0.094          \\ \midrule
\textbf{Full}        & 0.827         & 0.595         & 0.363          \\ \bottomrule
\end{tabular}
}
\end{table}

\mypara{Targeted Attack}
We first examine how attack performance changes under the \textit{targeted attack} scenario.
With the introduction of a database, the number of prior article samples available per candidate author increases substantially, up to 500 articles per candidate.
This is feasible because the article web search step can be bypassed, and all previous samples are pre-analyzed and aggregated into a single entry.
Consequently, the system obtains a more comprehensive representation of each author's profile while reducing the number of pairwise comparisons required.

\autoref{table:db_target} reports the F1 scores of the agent's predictions on whether a given article is written by the targeted author.
The increase in sample number significantly boosts prediction performance for both the LDA and SALA methods, as both benefit from richer author profiles.
With more samples available and analyzed individually, LDA also shows improved stability and reduced hallucination.
Our proposed SALA method benefits greatly from the expanded dataset, achieving an F1 score as high as 0.827.
The large number of reference articles allows the computed stylometric features to become more representative of each author's unique writing style.

The correlation between inference performance and the number of samples per candidate is further illustrated in \autoref{figure:targeted}.
Performance improves as more samples are incorporated, but the effect plateaus around 300 samples across varying numbers of target authors.
This suggests that while additional data enhances inference, a moderate number of samples already yields optimal performance.

With sufficiently strong authorship inference performance, we next analyze finer-grained feature contributions.
For the SALA method, we investigate which feature categories are most influential for authorship prediction.
\autoref{table:db_sala} presents the targeted attack results when SALA utilizes only one feature type at a time.
No single category dominates overall performance; however, lexical, syntactic, and style features contribute noticeably more than readability or semantic features.
This can be expected, given that our dataset consists of news articles, readability and sentiment-related features tend to be homogeneous across authors. 

\begin{table}[!t]
\centering
\caption{Percentage of correct inference with database module augmented in the open-world attack scenario.}
\label{table:db_open}
\scalebox{1.0}{
\begin{tabular}{@{}l|cc|cc@{}}
\toprule
\multirow{2}{*}{} & \multicolumn{2}{c|}{\textbf{Top-1}} & \multicolumn{2}{c}{\textbf{Top-3}} \\ \cmidrule(l){2-5} 
                  & \textbf{Full}  & \textbf{Filtered}  & \textbf{Full}  & \textbf{Filtered} \\ \midrule
\textbf{ES}       & 0.035          & 0.051              & 0.077          & 0.112             \\
\textbf{LDA}      & 0.042          & 0.056              & 0.092          & 0.123             \\
\textbf{SALA}     & 0.078          & 0.125              & 0.156          & 0.224             \\ \bottomrule
\end{tabular}
}
\end{table}

\begin{figure}[!t]
\centering
\includegraphics[width=\columnwidth]{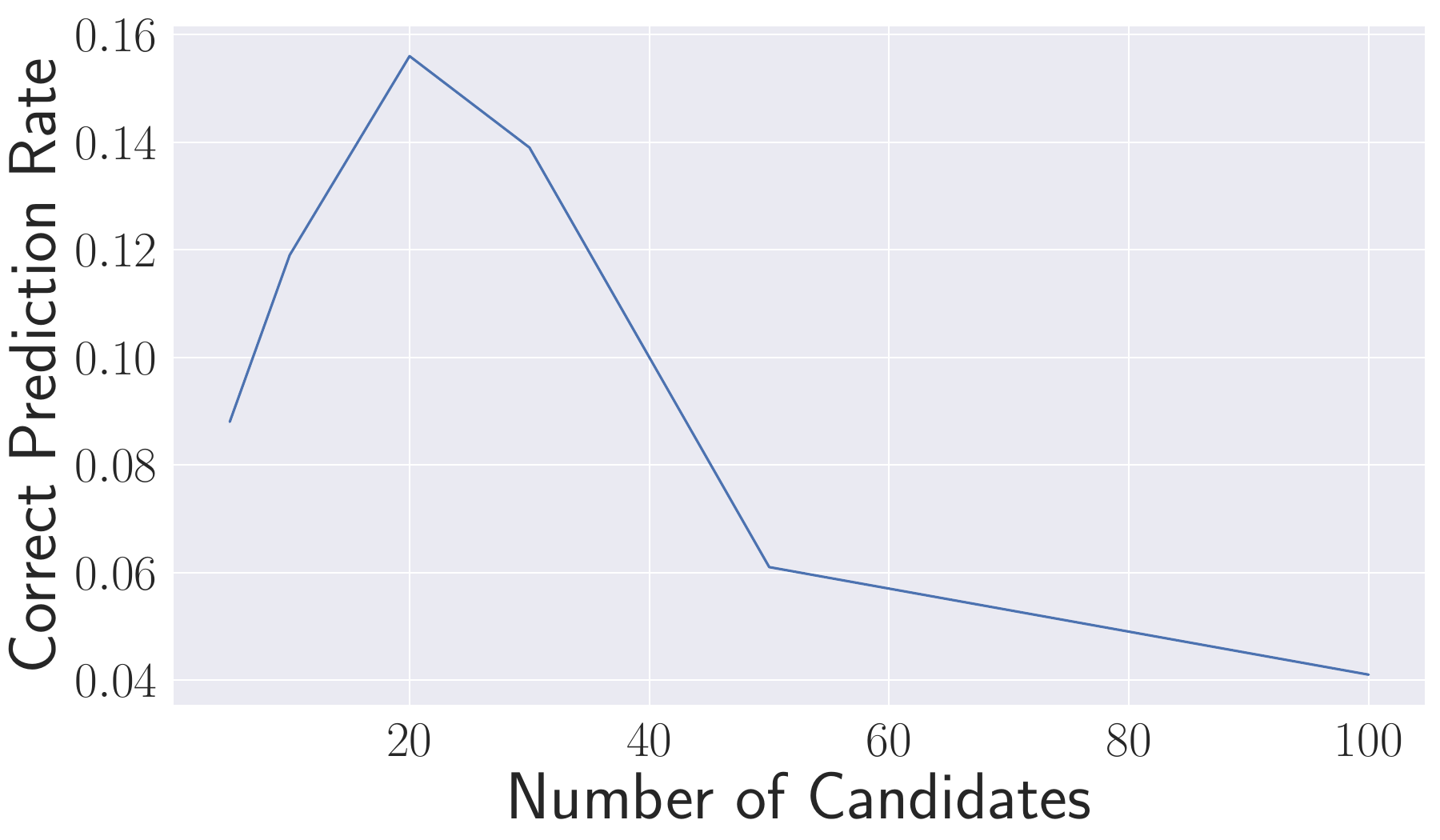}
\caption{Correct prediction rate of authorship inference (top-3) with database module augmented in the open-world scenario with different numbers of candidates selected during the search stage.}
\label{figure:openworld}
\end{figure}

\mypara{Open-World Attack}
We now examine the more challenging \textit{open-world} attack scenario with the database module enabled.
\autoref{table:db_open} shows that even when using the complete automated pipeline (i.e., search followed by matching), the agent augmented with the database can still make correct predictions, particularly when employing our proposed SALA matching strategy.
In addition to the increased efficacy of both SALA and LDA with larger numbers of available samples (as demonstrated previously), the substantial improvement in search performance ensures that enhanced matching accuracy is not offset by incorrect initial candidate selection.
Although the overall prediction performance remains relatively low, this is expected given the inherent difficulty of the open-world setting.

Unlike the targeted attack scenario, the number of candidate authors also affects the attack performance.
As the number of candidates increases, matching becomes more challenging since selecting the correct author from a larger pool is less likely.
Conversely, overly restricting the candidate set may degrade performance if the true author is omitted during the search stage.
As illustrated in \autoref{figure:openworld}, inference performance peaks when the number of candidates is set at a balanced level, large enough to include the correct author, yet small enough to maintain discriminative power.

Overall, the improved inference accuracy observed across both targeted and open-world settings highlights the significant deanonymization risks, particularly when augmented with a database module that aggregates and pre-processes extensive author profiles.

\section{Anonymization Enhancement}
\label{section:defense}

Given the effective inference capabilities demonstrated in the previous section, it is crucial to develop appropriate countermeasures to mitigate the associated deanonymization risks.
Most commercially available LLMs are equipped with built-in guardrails designed to prevent harmful or policy-violating behavior.
However, preventing deanonymization is particularly challenging due to the inherently ambiguous nature of the task.

During our development of the agent, we observed that only explicit and overt requests for deanonymization (e.g., ``Deanonymize the author of this article'') are typically rejected by the LLM's safety layer.
In contrast, if the request is reworded slightly (e.g., ``Compare this article's style to known authors''), the model will often proceed with the task.
This observation underscores the difficulty of enforcing strict prevention, as malicious intents can be easily disguised through minor prompt editing, even without employing sophisticated adversarial techniques such as jailbreaking~\cite{LGFXS23,DLLWZLWZL23,YLYX23,CRDHPW23,LXCX23,HGXLC23,WWLMW23}.

On one hand, our agent demonstrates the potential threat of adversarial usage, highlighting the need for more robust safeguards.
On the other hand, the agent's internal reasoning process provides valuable insight into how such deanonymization inferences are made, offering a foundation for reverse-engineering effective mitigation strategies.

To evaluate the effectiveness of our mitigation method introduced in \autoref{subsection:defense_module}, we conduct experiments under two challenging settings: (1) \textit{targeted} and (2) \textit{open-world} deanonymization attacks. We further compare our guided recomposition approach with baseline methods such as direct paraphrasing, in which an LLM is instructed to paraphrase the article without explicit guidance. To ensure the utility of the rewritten text, we compute embedding-based similarity scores between the original and rewritten articles, verifying that their semantic content remains largely unchanged.

\autoref{table:defense} reports the defense performance across scenarios. Our proposed method substantially reduces the success rate of deanonymization attacks, achieving near-random performance in the targeted setting and only marginal inference accuracy (approximately one percent) in the open-world setting. In contrast, simple paraphrasing proves less effective, as it often preserves sentence structure and high-level phrasing, leaving many author-revealing features intact.

Overall, by leveraging the \emph{Result Reflection Stage} to inform the \emph{guided recomposition}, our agent offers a practical and interpretable pathway for mitigating deanonymization risks posed by LLM-agent-powered inference systems.

\begin{table}[!t]
\centering
\caption{Authorship inference effectiveness with mitigation deployed.}
\label{table:defense}
\scalebox{1.0}{
\begin{tabular}{@{}lcc@{}}
\toprule
                        & \textbf{Targeted} & \textbf{Open-World} \\ \midrule
\textbf{No Defense}     & 0.827                                 & 0.156                              \\
\textbf{Direct Paraphrase} & 0.762                                 & 0.035  
\\
\textbf{Guided Recompose} & 0.561                                 & 0.012\\ \bottomrule
\end{tabular}
}
\end{table}

\section{Conclusion}
\label{section:conclusion}

We introduced a deanonymization agent for analyzing authorship risks in news articles, featuring a four-stage pipeline of information extraction, candidate search, candidate matching, and result analysis with anonymization suggestions.
Our proposed \textsc{SALA} (Stylometry-Assisted LLM Analysis) method combines stylometric features with LLM reasoning to achieve interpretable and accurate authorship inference, outperforming baseline approaches such as LDA and embedding similarity. 
Database augmentation further enhances both efficiency and performance, particularly in targeted and open-world attack scenarios.
Finally, through guided recomposition, the agent provides actionable anonymization strategies that effectively mitigate deanonymization risk while preserving content utility.

\section*{Limitations}
\label{section:limitations}

While our proposed framework demonstrates strong performance and interpretability, several limitations remain. First, the current evaluation is restricted to English-language news datasets, which may limit generalizability across domains, genres, or languages. Future work should explore more diverse and variable datasets, including social media, academic writing, and multilingual corpora, to better assess the robustness of the approach. Additionally, although the guided recomposition strategy effectively reduces identifiability, its rewriting quality and stylistic preservation could be further improved through adaptive prompting or human-in-the-loop refinement. Finally, while the database module enhances inference performance, it also introduces potential bias from uneven author representation, suggesting the need for more balanced or dynamically updated author profiles.

\section*{Ethics Statement}
\label{section:ethics}

This work aims to raise awareness of potential deanonymization risks posed by LLM-based authorship analysis and to promote the development of responsible mitigation strategies. By presenting this framework, we hope to facilitate future research on mitigating potential risks of LLM agents in real-world applications.

\begin{small}
\balance
\bibliographystyle{plain}
\bibliography{normal_generated_py3}
\end{small}

\appendix

\end{document}